# Non-Deterministic Policies in
# Markovian Decision Processes

**Mahdi Milani Fard**                                    MMILAN1@CS.MCGILL.CA
**Joelle Pineau**                                        JPINEAU@CS.MCGILL.CA
*Reasoning and Learning Laboratory*
*School of Computer Science, McGill University*
*Montreal, QC, Canada*

## Abstract

Markovian processes have long been used to model stochastic environments. Reinforcement learning has emerged as a framework to solve sequential planning and decision-making problems in such environments. In recent years, attempts were made to apply methods from reinforcement learning to construct decision support systems for action selection in Markovian environments. Although conventional methods in reinforcement learning have proved to be useful in problems concerning sequential decision-making, they cannot be applied in their current form to decision support systems, such as those in medical domains, as they suggest policies that are often highly prescriptive and leave little room for the user's input. Without the ability to provide flexible guidelines, it is unlikely that these methods can gain ground with users of such systems.

This paper introduces the new concept of non-deterministic policies to allow more flexibility in the user's decision-making process, while constraining decisions to remain near optimal solutions. We provide two algorithms to compute non-deterministic policies in discrete domains. We study the output and running time of these method on a set of synthetic and real-world problems. In an experiment with human subjects, we show that humans assisted by hints based on non-deterministic policies outperform both human-only and computer-only agents in a web navigation task.

## 1. Introduction

Planning and decision-making have been well studied in the AI community. Intelligent agents have been designed and developed to act in, and interact with, a variety of environments. This usually involves sensing the environment, making a decision using some intelligent inference mechanism, and then performing an action on the environment (Russell & Norvig, 2003). Often times, this process involves some level of learning, along with the decision-making process, to make the agent more efficient in performing the intended goal.

Reinforcement Learning (RL) is a branch of AI that tries to develop a computational approach to solving the problem of learning through interaction. RL is the process of learning what to do—how to map situations to actions—so as to maximize a numerical reward signal (Sutton & Barto, 1998). Many methods have been developed to solve the RL problem with different types of environments and different types of agents. However, most of the work in RL has focused on autonomous agents such as robots or software agents. The RL controllers are thus designed to issue a single action at each time-step which will be executed by the acting agent. In the past few years, methods developed by the RL community have started to be used in sequential decision support systems (Murphy, 2005; Pineau, Bellemare, Rush, Ghizaru, & Murphy, 2007; Thapa, Jung, & Wang, 2005; Hauskrecht & Fraser, 2000). In many of these systems, a human being makes the final decision. Usability and acceptance issues thus become important in these cases. Most RL methods therefore require some level of adaptation to be used with decision support systems. Such adaptations are the main contribution of this paper.

Medical domains are among the cases for which RL needs further adaptation. Although the RL framework correctly models sequential decision-making in complex medical scenarios, including long-term treatment design, standard RL methods cannot be applied to medical settings in their current form as they lack flexibility in their suggestions. Such requirements are, of course, not specific to medical domains and, for instance, might be needed in an aircraft controller that provides suggestions to a pilot.

The important difference between decision support system and the classical RL problem stems from the fact that in the decision support system, the acting agent is often a human being, which of course has his/her own decision process. Therefore, the assumption that the controller should only send one clear commanding signal to the acting agent is not appropriate. It is more accurate to assume that some aspect of the decision-making process will be influenced by the user of the system.

This view of the decision process is particularly relevant in two different situations. First, in many practical cases, we do not have an exact model of the system. Instead, we may have a noisy model built from a finite number of interactions with the environment. This leads to a type of uncertainty that is usually referred to as *extrinsic uncertainty*. Most RL algorithms ignore this uncertainty and assume that the model is perfect. However if we look closely, the performance of the optimal action based on an imperfect model might not be statistically different from the next best action. Bayesian approaches have looked at this problem by providing confidence measure over the agent's performance (Mannor, Simester, Sun, & Tsitsiklis, 2007). In cases where the acting agent is a human being, we can use these confidence measures to provide the user with a complete set of actions, each of which might be the optimal and for which we do not have enough evidence to differentiate. The user can then use his/her expertise to make the final decision. Such methods should guarantee that the suggestions provided by the system are statistically meaningful and plausible.

On the other hand, even when we do have complete knowledge of the system and we can identify the optimal action, there might still be other actions which are "roughly equal" in performance. At this point, the decision between these near-optimal options could be left to the acting agent—namely the human being that is using the decision support system. This could have many advantages, ranging from better user experience, to increased robustness and flexibly. Among the near-optimal solutions, the user can select based on further domain knowledge, or any other preferences, that are not captured by the system. For instance, in a medical diagnosis system that suggests treatments, providing the physician with several options might be useful as the final decision could be made based on further knowledge of the patient's medical status, or preferences regarding side effects.

Throughout this paper we address the latter issue through a combination of theoretical and empirical investigations. We introduce the new concept of non-deterministic policies to capture a decision-making process intended for decision support systems. Such policies involve suggesting a set of actions, from which a non-deterministic choice is made by the user. We apply this formulation to solve the problem of finding near-optimal policies to provide flexible suggestions to the user.

In particular, we investigate how we can suggest several actions to the acting agent, while providing performance guarantees with a worst-case analysis. Section 2 introduces the necessary technical background material. Section 3 defines the concept of non-deterministic policies and related concepts. Section 4 addresses the problem of providing choice to the acting agent while keeping near-optimality guarantees on the performance of the worst-case scenario. We propose two algorithms to solve these problems and provide approximation techniques to speed up the computation for larger domains.

Methods introduced in this paper are general enough to apply to any decision support system in an observable Markovian environment. Our empirical investigations focus primarily on sequential decision-making problems in clinical domains, where the system should provide suggestions on the best treatment options for patients. These decisions are provided over a sequence of treatment phases. These systems are specifically interesting because often times, different treatment options seem to provide only slightly different results. Therefore, providing the physician with several suggestions would be beneficial in improving the usability of the system and the performance of the final decision.

## 2. Definitions and Notations

This section introduces the main notions behind sequential decision-making and the mathematical formulations used in RL.

### 2.1 Markov Decision Processes

A Markov Decision Process (MDP) is a model of system dynamics in sequential decision problems that involves probabilistic uncertainty about future states of the system (Bellman, 1957). MDPs are used to model the interactions between an agent and an observable Markovian environment. The system is assumed to be in a state at any given time. The agent observes the state and performs an action accordingly. The system then makes a transition to the next state and the agent receives some reward.

Formally, an MDP is defined by the 5-tuple $(S, A, T, R, \gamma)$:

- **States**: $S$ is the set of states. The state usually captures the complete configuration of the system. Once the state of the system is known, the future of the system is independent from all previous system transitions. This means that the state of the system is a sufficient statistic of the history of the system.

- **Actions**: $A : S \to 2^{\mathcal{A}}$ is the set of actions allowed in each state where $\mathcal{A}$ is the set of all actions. $A(s)$ is the set of actions the agent can choose from, while interacting with the system in state $s$.

- **Transition Probabilities**: $T : S \times \mathcal{A} \times S \to [0, 1]$ defines the transition probabilities of the system. This function specifies how likely it is to end up at any state, given the current state and a specific action performed by the agent. Transition probabilities are specified based on the Markovian assumption. That is, if the state of the system at time $t$ is denoted by $s_t$ and the action at that time is $a_t$, then we have:

$$\Pr(s_{t+1}|a_t, s_t, a_{t-1}, a_{t-1}, \ldots, a_0, s_0) = Pr(s_{t+1}|a_t, s_t). \tag{1}$$

We focus on *homogeneous* processes in which the system dynamics are independent of the time. Thus the transition function is stationary with respect to time:

$$T(s, a, s') \stackrel{\text{def}}{=} Pr(s_{t+1} = s'|a_t = a, s_t = s). \tag{2}$$

- **Rewards**: $R : S \times \mathcal{A} \times \mathbb{R} \to [0, 1]$ is the probabilistic reward model. Depending on the current state of the system and the action taken, the agent will receive a reward drawn from this model. We focus on homogeneous processes in which, again, the reward distribution does not change over time. If the reward at time $t$ is denoted by $r_t$, then we have:

$$r_t \sim R(s_t, a_t). \tag{3}$$

  Depending on the domain, the reward could be deterministic or stochastic. We use the general stochastic model throughout this paper. The mean of this distribution is denoted by $\bar{R}(s, a)$.

- **Discount Factor**: $\gamma \in [0, 1)$ is the discount rate used to calculate the long-term return.

The agent starts in an initial state $s_0 \in S$. At each time step $t$, an action $a_t \in A(s_t)$ is taken by the agent. The system then makes a transition to $s_{t+1} \sim T(s_t, a_t)$ and the agent receives an immediate reward $r_t \sim R(s_t, a_t)$.

The goal of the agent is to maximize the discounted sum of rewards over the planning horizon $h$ (could be infinite). This is usually referred to as the *return* (denoted by $D$):

$$D = \sum_{t=0}^{h} \gamma^t r_t. \tag{4}$$

In the finite horizon case, this sum is taken up to the horizon limit and the discount factor can be set to 1. However, in the infinite horizon case the discount factor should be less than 1 so that the return has finite value. The return on the process depends on both the stochastic transitions and rewards, as well as the actions taken by the agent.

Often times the transition structure of the MDP contains no loop with non-zero probability. Such a transition graph can be modeled by a directed acyclic graph (DAG). This class of MDPs is interesting as it includes multi-step decision-making in finite horizons, such as those found in medical domains.

## 2.2 Policy and Value Function

A policy is a way of defining the agent's action selection with respect to the changes in the environment. A (probabilistic) policy on an MDP is a mapping from the state space to a distribution over the action space:

$$\pi : S \times \mathcal{A} \to [0, 1]. \tag{5}$$

A deterministic policy is a policy that defines a single action per state. That is, $\pi(s) \in A(s)$. We will later introduce the notion of non-deterministic policies on MDPs to deal with sets of actions.

The agent interacts with the environment and takes actions according to the policy. The value function of the policy is defined to be the expectation of the return given that the agent acts according to that policy:

$$V^\pi(s) \stackrel{\text{def}}{=} \mathbb{E}[D^\pi(s)] = \mathbb{E}\left[\sum_{t=0}^{\infty} \gamma^t r_t | s_0 = s, a_t = \pi(s_t)\right]. \tag{6}$$

Using the linearity of the expectation, we can write the above expression in a recursive form, known as the *Bellman equation* (Bellman, 1957):

$$V^\pi(s) = \sum_{a \in A} \pi(s, a) \left[\bar{R}(s, a) + \sum_{s' \in S} T(s, a, s') V^\pi(s')\right]. \tag{7}$$

The value function has been used as the primary measure of performance in much of the RL literature. There are, however, some ideas that take the risk or the variance of the return into account as a measure of optimality (Heger, 1994; Sato & Kobayashi, 2000). The more common criteria, though, is to assume that the agent is trying to find a policy that maximizes the value function. Such a policy is referred to as the optimal policy.

We can also define the value function over the state-action pairs. This is usually referred to as the $Q$-function, or the $Q$-value, of that pair. By definition:

$$Q^\pi(s, a) \stackrel{\text{def}}{=} \mathbb{E}[D^\pi(s, a)] = \mathbb{E}\left[\sum_{t=0}^{\infty} \gamma^t r_t | s_0 = s, a_0 = a, t \geq 1 : a_t = \pi(s_t)\right]. \tag{8}$$

That is, the $Q$-value is the expectation of the return, given that the agent starts in state $s$, takes action $a$, and then follows policy $\pi$. The $Q$-function also satisfies the Bellman equation:

$$Q^\pi(s, a) = \bar{R}(s, a) + \sum_{s' \in S} T(s, a, s') \sum_{a' \in A} \pi(s', a') Q^\pi(s', a'), \tag{9}$$

which can be rewritten as:

$$Q^\pi(s, a) = \bar{R}(s, a) + \sum_{s' \in S} T(s, a, s') V^\pi(s'). \tag{10}$$

The $Q$-function is often used to compare the optimality of different actions given a fixed subsequent policy.

## 2.3 Planning Algorithms and Optimality

The optimal policy, denoted by $\pi^*$, is defined to be the policy that maximizes the value function at the initial state:

$$\pi^* = \operatorname*{argmax}_{\pi \in \Pi} V^\pi(s_0). \tag{11}$$

It has been shown (Bellman, 1957) that for any MDP, there exists an optimal deterministic policy that is no worse than any other policy for that MDP. The value of the optimal policy $V^*$ satisfies the *Bellman optimality equation*:

$$V^*(s) = \max_{a \in A} \left[ \bar{R}(s,a) + \sum_{s' \in S} T(s,a,s')V^*(s') \right]. \tag{12}$$

The deterministic optimal policy follows from this:

$$\pi^*(s) = \operatorname*{argmax}_{a \in A} \left[ \bar{R}(s,a) + \sum_{s' \in S} T(s,a,s')V^*(s') \right]. \tag{13}$$

Alternatively we can write these equations with the $Q$-function:

$$Q^*(s,a) = \bar{R}(s,a) + \sum_{s' \in S} T(s,a,s')V^*(s'). \tag{14}$$

Thus $V^*(s) = \max_a Q^*(s,a)$ and $\pi^*(s) = \operatorname{argmax}_a Q^*(s,a)$.

Much of the literature in RL has focused on finding the optimal policy. There are many methods developed for policy optimization. One way to find the optimal policy is to solve the Bellman optimality equation and then use Eqn 13 to choose the actions. The Bellman optimality equation can be formulated as a simple linear program (Bertsekas, 1995):

$$\min_V \mu^T V, \text{ subject to}$$
$$V(s) \geq \bar{R}(s,a) + \gamma \sum_{s'} T(s,a,s')V(s') \quad \forall s,a, \tag{15}$$

where $\mu$ represents an initial distribution over the states. The solution to the above problem is the optimal value function. Notice that $V$ is represented in matrix form in this equation. It is known that linear programs can be solved in polynomial time (Karmarkar, 1984). However, solving them might become impractical in large (or infinite) state spaces. Therefore often times methods based on dynamic programming are preferred to the linear programming solution.

## 3. Non-Deterministic Policies: Definition and Motivation

We begin this section by considering the problem of decision-making in sequential decision support systems. Recently, MDPs have emerged as useful frameworks for optimizing action choices in the context of medical decision support systems (Schaefer, Bailey, Shechter, & Roberts, 2004; Hauskrecht & Fraser, 2000; Magni, Quaglini, Marchetti, & Barosi, 2000; Ernst, Stan, Concalves, & Wehenkel, 2006). Given an adequate MDP model (or data source), many methods can be used to find a good action-selection policy. This policy is usually a *deterministic* or *stochastic* function. But policies of these types face a substantial barrier in terms of gaining acceptance from the medical community, because they are highly prescriptive and leave little room for the doctor's input. These problems are, of course, not specific to the medical domain and are present in any application where the actions are executed by a human. In such cases, it may be preferable to provide several equivalently good action choices, so that the agent can pick among those according to his or her own heuristics and preferences.

To address this problem, this work introduces the notion of a *non-deterministic policy*, which is a function mapping each state to a set of actions, from which the acting agent can choose.

**Definition 1.** *A **non-deterministic policy** $\Pi$ on an MDP $(S, A, T, R, \gamma)$ is a function that maps each state $s \in S$ to a non-empty set of actions denoted by $\Pi(s) \subseteq A(s)$. The agent can choose to do any action $a \in \Pi(s)$ whenever the MDP is in state $s$.*

**Definition 2.** *The **size** of a non-deterministic policy $\Pi$, denoted by $|\Pi|$, is the sum of the cardinality of the action sets in $\Pi$: $|\Pi| = \sum_s |\Pi(s)|$.*

In the following sections we discuss two scenarios in which non-deterministic policies can be useful. We show how they can be used to implement more robust decision support systems with statistical guarantees of performance.

## 3.1 Providing Choice to the Acting Agent

Even in cases where we have complete knowledge of the dynamics of the planning problem at hand, and when we can accurately calculate actions' utilities, it might not be desirable to provide the user with only the optimal choice of action at each time step. In some domains, the difference between the utility of the top few actions may not be substantial. In medical decision-making, for instance, this difference may not be *medically significant* based on the given state variables.

In such cases, it seems natural to let the user decide between the top few actions, using his/her own expertise in the domain. This results in a further injection of domain knowledge in the decision-making process, thus making it more robust and practical. Such decisions can be based on facts known to the user that are not incorporated in the automated planning system. It can also be based on preferences that might change from case to case. For instance, a doctor can get several recommendations as to how to treat a patient to maximize the chance of remission, but then decide what medication to apply considering also the patient's medical record, preferences regarding side effects, or medical expenses.

This idea of providing *choice* to the user should be accompanied by reasonable guarantees on the performance of the final decision, regardless of the choice made by the user. A notion of near-optimality should be enforced to make sure the actions are never far from the best possible option. Such guarantees are enforced by providing a worst-case analysis on the decision process.

## 3.2 Handling Model Uncertainty

In many practical cases we do not have complete knowledge of the system at hand. Instead, we may get a set of trajectories collected from the system according to some specific policy. In some cases, we may be given the chance to choose this policy (in on-line and active RL), and in other cases we may have access only to data from some fixed policy. In medical trials, in particular, data is usually collected according to a randomized policy, fixed ahead of time through consultation with the clinical researchers.

Given a set of sample trajectories, we can either build a model of the domain (in model-based approaches) or directly estimate the utility of different actions (with model-free approaches). However these models and estimates are not always accurate when we only observe a finite amount of data. In many cases, the data may be too sparse and incomplete to uniquely identify the best option. That is, the difference in the performance measure of different actions is not statistically significant.

There are other cases where it might be useful to let the user decide on the final choice between those actions for which we do not have enough evidence to differentiate. This comes with the assumption that the user can identify the best choice among those that are recommended. The task is therefore to provide the user with a small set of actions that will almost surely include the optimal one.

In this paper we only focus on the problem of providing flexible policies with near-optimal performance. Using non-deterministic policies for handling model uncertainty remains an interesting future work.

## 4. Near-Optimal Non-Deterministic Policies

Often times, it is beneficial to provide the user of a decision support system with a set of near-optimal solutions. With MDPs, this would be to suggest a set of near-optimal actions to the user and let the user make a decision among the proposed actions. The notion of near-optimality should therefore be on the set of all possible policies that are consistent with the proposed actions. That is, no matter which action is chosen among the proposed options at each state, the final performance should be close to that of the optimal policy. Such constraint suggests a worst-case analysis of the decision-making process. Therefore, we opt to guarantee the performance of any action selection consistent with a non-deterministic policy by putting the near-optimality constraint on the worst-case selection of actions by the user.

**Definition 3.** *The **(worst-case) value of a state-action pair*** $(s, a)$ *according to a* non-deterministic policy $\Pi$ *on an MDP* $M = (S, A, T, R, \gamma)$ *is given by the recursive definition:*

$$Q_M^{\Pi}(s, a) = \bar{R}(s, a) + \gamma \sum_{s' \in S} \left( T(s, a, s') \min_{a' \in \Pi(s')} Q_M^{\Pi}(s', a') \right), \tag{16}$$

*which is the worst-case expected return under the allowed set of actions.*

**Definition 4.** *We define the **(worst-case) value of a state*** $s$ *according to a non-deterministic policy* $\Pi$*, denoted by* $V_M^{\Pi}(s)$*, to be:*

$$\min_{a \in \Pi(s)} Q_M^{\Pi}(s, a). \tag{17}$$

To calculate the value of a non-deterministic policy, we construct an *evaluation MDP*, $M' = (S, A', R', T, \gamma)$, where $A' = \Pi$ and $R' = -R$.

**Theorem 1.** *The negated value of the non-deterministic policy* $\Pi$ *is equal to that of the optimal policy on the evaluation MDP:*

$$Q_M^{\Pi}(s, a) = -Q_{M'}^{*}(s, a). \tag{18}$$

*Proof.* We show that if $Q^*_{M'}$ satisfies the Bellman optimality equation on $M'$, then the negated values satisfy Eqn 16 on $M$:

$$Q^*_{M'}(s,a) = \bar{R}'(s,a) + \gamma \sum_{s' \in S} T(s,a,s') \max_{a' \in A'} Q^*_{M'}(s',a') \qquad (19)$$

$$\Rightarrow -Q^*_{M'}(s,a) = -\bar{R}'(s,a) - \gamma \sum_{s' \in S} T(s,a,s') \max_{a' \in A'} Q^*_{M'}(s',a') \qquad (20)$$

$$\Rightarrow -Q^*_{M'}(s,a) = \bar{R}(s,a) + \gamma \sum_{s' \in S} T(s,a,s') \min_{a' \in \Pi(s')} -Q^*_{M'}(s',a'), \qquad (21)$$

which is equivalent to Eqn 16 for $Q^\Pi_M(s,a) = -Q^*_{M'}(s,a)$. □

This means that policy evaluation for a non-deterministic policy can be achieved by any method that finds the optimal policy on an MDP.

**Definition 5.** *A non-deterministic policy $\Pi$ is said to be **augmented** with state-action pair $(s,a)$, denoted by $\Pi' = \Pi + (s,a)$, if it satisfies:*

$$\Pi'(s') = \begin{cases} \Pi(s'), & s' \neq s \\ \Pi(s') \cup \{a\}, & s' = s. \end{cases} \qquad (22)$$

If a policy $\Pi$ can be achieved by a number of augmentations from a policy $\Pi'$, we say that $\Pi$ includes $\Pi'$.

**Definition 6.** *A non-deterministic policy $\Pi$ is said to be **non-augmentable** according to a constraint $\Psi$ if and only if $\Pi$ satisfies $\Psi$, and for any state-action pair $(s,a)$, $\Pi + (s,a)$ does not satisfy $\Psi$.*

In this paper we will be working with constraints that have this particular property: if a policy $\Pi$ does not satisfy $\Psi$, any policy that includes $\Pi$ does not satisfy $\Psi$. We will refer to such constraints as being *monotonic*. One such constraint is $\epsilon$-optimality, which is discussed in the next section.

### 4.1 $\epsilon$-Optimal Non-Deterministic Policies

**Definition 7.** *A non-deterministic policy $\Pi$ on an MDP $M$ is said to be $\epsilon$-**optimal**, with $\epsilon \in [0,1]$, if we have[1]:*

$$V^\Pi_M(s) \geq (1-\epsilon)V^*_M(s), \quad \forall s \in S. \qquad (23)$$

This can be thought of as a constraint $\Psi$ on the space of non-deterministic policies, set to ensure that the worst-case expected return is within some range of the optimal value.

**Theorem 2.** *The $\epsilon$-optimality constraint is monotonic.*

---

1. In some of the MDP literature, $\epsilon$-optimality is defined as an additive constraint ($Q^\Pi_M \geq Q^*_M - \epsilon$) (Kearns & Singh, 2002). The derivations will be analogous in that case. We chose the multiplicative constraint as it has cleaner derivations.

*Proof.* Suppose $\Pi$ is not $\epsilon$-optimal. Then for any augmentation $\Pi' = \Pi + (s, a)$, we have:

$$
\begin{aligned}
Q_M^{\Pi'}(s, a) &= \bar{R}(s, a) + \gamma \sum_{s' \in S} \left( T(s, a, s') \min_{a' \in \Pi'(s')} Q_M^{\Pi'}(s', a') \right) \\
&\leq \bar{R}(s, a) + \gamma \sum_{s' \in S} \left( T(s, a, s') \min_{a' \in \Pi(s')} Q_M^{\Pi'}(s', a') \right) \\
&\leq Q_M^{\Pi}(s, a),
\end{aligned}
$$

which implies:

$$
V_M^{\Pi'}(s) \leq V_M^{\Pi}(s).
$$

As $\Pi$ was not $\epsilon$-optimal, this means that $\Pi'$ will not be $\epsilon$-optimal either as the value function can only decrease with the policy augmentation. $\square$

More intuitively, it follows from the fact that adding more options cannot increase the minimum utility as the former worst case choice is still available after the augmentation.

**Definition 8.** *A **conservative** $\epsilon$-**optimal** non-deterministic policy $\Pi$ on an MDP $M$ is a policy that is non-augmentable according to the following constraint:*

$$
\bar{R}(s, a) + \gamma \sum_{s'} \left( T(s, a, s')(1 - \epsilon) V_M^*(s') \right) \geq (1 - \epsilon) V_M^*(s), \quad \forall a \in \Pi(s). \tag{24}
$$

This constraint indicates that we only add those actions to the policy whose reward plus $(1 - \epsilon)$ of the future optimal return is within the sub-optimal margin. This ensures that the non-deterministic policy is $\epsilon$-optimal by using the inequality:

$$
Q_M^{\Pi}(s, a) \geq \bar{R}(s, a) + \gamma \sum_{s'} \left( T(s, a, s')(1 - \epsilon) V_M^*(s') \right), \tag{25}
$$

instead of solving Eqn 16 and using the inequality constraint in Eqn 23. Applying Eqn 24 guarantees that the non-deterministic policy is $\epsilon$-optimal while it may still be augmentable according to Eqn 23, hence the name conservative.

It can also be shown that the conservative policy is unique. This is because if there were two different conservative policies, then the union of them would be conservative, which violates the assumption that they are non-augmentable according to Eqn 24.

**Definition 9.** *A **non-augmentable** $\epsilon$-**optimal** non-deterministic policy $\Pi$ on an MDP $M$ is a policy that is non-augmentable according to the constraint in Eqn 23.*

This is a non-deterministic policy for which adding more actions violates the near-optimality constraint of the worst-case performance. In a search for $\epsilon$-optimal policies, a non-augmentable one has a *locally maximal size*. This means that although the policy might not be the largest among the $\epsilon$-optimal policies, we cannot add any more actions to it without removing other actions, hence the locally maximal reference.

Any non-augmentable $\epsilon$-optimal policy includes the conservative policy. This is because we can always add the conservative policy to any policy and remain within the $\epsilon$ bound.

However, non-augmentable $\epsilon$-optimal policies are not necessarily unique, as they have only locally maximal size.

In the remainder of this section, we focus on the problem of searching over the space of non-augmentable $\epsilon$-optimal policies, such as to maximize some criteria. Specifically, we aim to find non-deterministic policies that give the acting agent more options while staying within an acceptable sub-optimal margin.

We now present an example that clarifies the concepts introduced so far. To simplify the presentation of the example, we assume deterministic transitions. However, the concepts apply as well to any probabilistic MDP. Figure 1 shows an example MDP. The labels on the arcs show action names and the corresponding rewards are shown in the parentheses. We assume $\gamma \simeq 1$ and $\epsilon = 0.05$. Figure 2 shows the optimal policy of this MDP. The conservative $\epsilon$-optimal non-deterministic policy of this MDP is shown in Figure 3.

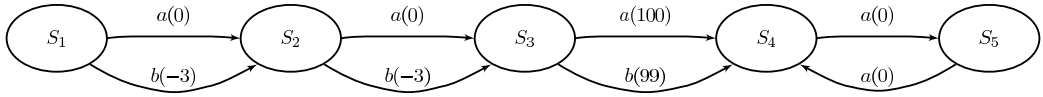

Figure 1: Example MDP

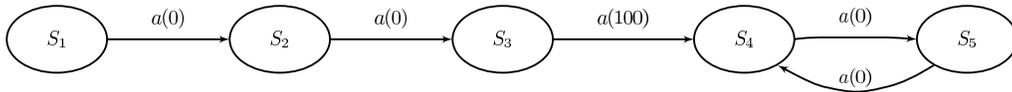

Figure 2: Optimal policy

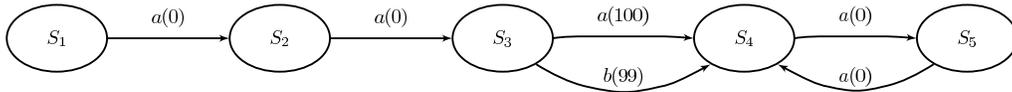

Figure 3: Conservative $\epsilon$-optimal policy

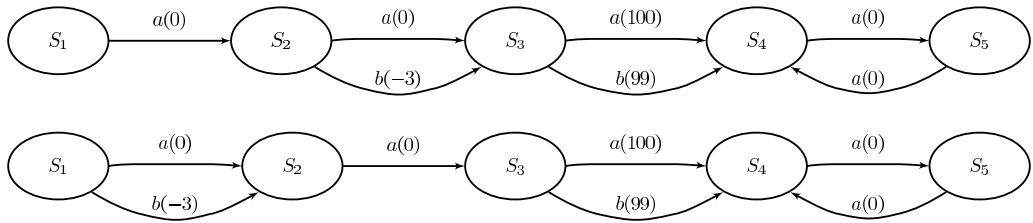

Figure 4: Two non-augmentable $\epsilon$-optimal policies

Figure 4 includes two possible non-augmentable $\epsilon$-optimal policies. Although both policies in Figure 4 are $\epsilon$-optimal, the union of these is not $\epsilon$-optimal. This is due to the fact that adding an option to one of the states removes the possibility of adding options to other states, which illustrates why local changes to the policy are not always appropriate when searching in the space of $\epsilon$-optimal policies.

## 4.2 Optimization Criteria

We formalize the problem of finding an $\epsilon$-optimal non-deterministic policy in terms of an optimization problem. There are several optimization criteria that can be formulated, while still complying with the $\epsilon$-optimality constraint.

- **Maximizing the size of the policy**: According to this criterion, we seek non-augmentable $\epsilon$-optimal policies that have the biggest overall size (Def 2). This provides more options to the agent while still keeping the $\epsilon$-optimal guarantees. The algorithms proposed in later sections use this optimization criterion. Notice that the solution to this optimization problem is non-augmentable according to the $\epsilon$-optimal constraint, because it maximizes the overall size of the policy.

  As a variant of this, we can try to maximize the sum of the log of the size of the action sets:

  $$\sum_{s \in S} \log |\Pi(s)|. \tag{26}$$

  This enforces a more even distribution of choice on the action set. However, we will be using the basic case of maximizing the overall size as it will be an easier optimization problem.

- **Maximizing the margin**: We can aim to maximize the *margin of a non-deterministic policy* $\Pi$:

  $$\max_{\Pi} \Phi_M(\Pi), \tag{27}$$

  where:

  $$\Phi_M(\Pi) = \min_{s \in S} \left( \min_{a \in \Pi(s), a' \notin \Pi(s)} \left( Q(s, a) - Q(s, a') \right) \right). \tag{28}$$

  This optimization criterion is useful when one wants to find a clear separation between the good and bad actions in each state.

- **Minimizing the uncertainty**: If we learn the models from data we will have some uncertainty about the optimal action in each state. We can use some variance estimation on the value function (Mannor, Simester, Sun, & Tsitsiklis, 2004) along with a Z-Test to get some confidence level on our comparisons and find the probability of having the wrong order when comparing actions according to their values. Let $Q$ be the value of the true model and $\hat{Q}$ be our empirical estimate based on some dataset $D$. We aim to minimize the *uncertainty of a non-deterministic policy* $\Pi$:

  $$\min_{\Pi} \Phi_M(\Pi), \tag{29}$$

  where:

  $$\Phi_M(\Pi) = \max_{s \in S} \left( \max_{a \in \Pi(s), a' \notin \Pi(s)} Pr \left\{ Q(s, a) < Q(s, a') | D \right\} \right). \tag{30}$$

Notice that the last two criteria can be defined both in the space of all $\epsilon$-optimal policies, or only the non-augmentable ones.

In the following sections we provide algorithms to solve the first optimization problem mentioned above, which aims to maximize the size of the policy. We focus on this criterion as it seems most appropriate for medical decision support systems, where it is desirable for the acceptability of the system to find policies that provide as much choice as possible for the acting agent. Developing algorithms to address the other two optimization criteria remains an interesting open problem.

### 4.3 Maximal $\epsilon$-Optimal Policy

The exact computational complexity of finding a maximal $\epsilon$-optimal policy is not yet known. The problem is certainly NP, as one can find the value of the non-deterministic policy in polynomial time by solving the evaluation MDP with a linear program. We suspect that the problem is NP-complete, but we have yet to find a reduction from a known NP-complete problem.

In order to find the largest $\epsilon$-optimal policy, we present two algorithms. We first present a Mixed Integer Program (MIP) formulation of the problem, and then present a search algorithm that uses the monotonic property of the $\epsilon$-optimal constraint. While the MIP method is useful as a general and theoretical formulation of the problem, the search algorithm has potential for further extensions with heuristics.

#### 4.3.1 Mixed Integer Programming Solution

Recall that we can formulate the problem of finding the optimal deterministic policy of an MDP as a simple linear program (Bertsekas, 1995):

$$\min_V \mu^T V, \text{ subject to}$$
$$V(s) \geq \bar{R}(s,a) + \gamma \sum_{s'} T(s,a,s')V(s') \quad \forall s,a, \tag{31}$$

where $\mu$ can be thought of as the initial distribution over states. The solution to the above problem is the optimal value function ($V^*$). Similarly, having computed $V^*$ using Eqn 31, the problem of searching for an optimal non-deterministic policy according to the size criterion can be rewritten as a Mixed Integer Program:[2]

$$\max_{V,\Pi}(\mu^T V + (V_{max} - V_{min})e_s^T \Pi e_a), \text{ subject to}$$
$$V(s) \geq (1-\epsilon)V^*(s) \qquad \forall s$$
$$\sum_a \Pi(s,a) > 0 \qquad \forall s$$
$$V(s) \leq \bar{R}(s,a) + \gamma \sum_{s'} T(s,a,s')V(s') + V_{max}(1 - \Pi(s,a)) \quad \forall s,a. \tag{32}$$

Here we are overloading the notation $\Pi$ to define a binary matrix representing the policy, where $\Pi(s,a)$ is 1 if $a \in \Pi(s)$, and 0 otherwise. We define $V_{max} = R_{max}/(1-\gamma)$ and $V_{min} = R_{min}/(1-\gamma)$. The $e$'s are column vectors of 1 with the appropriate dimensions. The first set of constraints makes sure that we stay within $\epsilon$ of the optimal return. The second set of constraints ensures that at least one action is selected per state. The third set ensures that for those state-action pairs that are chosen in any policy, the Bellman constraint holds, and otherwise, the constant $V_{max}$ makes the constraint trivial. Notice that the solution to the above problem maximizes $|\Pi|$ and the result is non-augmentable.

---

2. Note that in this MIP, unlike the standard LP for MDPs, the choice of $\mu$ can affect the solution in cases where there is a tie in the size of $\Pi$.

**Theorem 3.** *The solution to the mixed integer program of Eqn 32 is non-augmentable according to $\epsilon$-optimality constraint.*

*Proof.* First, notice that the solution is $\epsilon$-optimal, due to the first set of constraints on the (worst-case) value function. To show that it is non-augmentable, as a counter argument, suppose that we could add a state-action pair to the solution $\Pi$, while still staying in $\epsilon$ sub-optimal margin. By adding that pair, the objective function is increased by $(V_{max} - V_{min})$, which is bigger than any possible decrease in the $\mu^T V$ term, and thus the objective is improved, which conflicts with $\Pi$ being the solution. $\qquad\square$

We can use any MIP solver to solve the above problem. Note however that we do not make use of the monotonic nature of the constraints. A general purpose MIP solver could end up searching in the space of all the possible non-deterministic policies, which would require a running time exponential in the number of state-action pairs ($O(2^{|S||A|+\delta})$).

#### 4.3.2 HEURISTIC SEARCH

Alternatively, we develop a heuristic search algorithm to find a maximal $\epsilon$-optimal policy. We can make use of the monotonic property of the $\epsilon$-optimal policies to narrow down the search. We start by computing the conservative policy. We then augment it until we arrive at a non-augmentable policy. We also make use of the fact that if a policy is not $\epsilon$-optimal, neither is any other policy that includes it, and thus we can cut the search tree at this point.

Table 1: Heuristic search algorithm to find $\epsilon$-optimal policies with maximum size

---

**Function** $getOptimal(\Pi, startIndex, \epsilon)$
$\Pi_o \leftarrow \Pi$
**for** $i \leftarrow startIndex$ **to** $|S||A|$ **do**
    $(s, a) \leftarrow p_i$
    **if** $a \notin \Pi(s)$ & $V(\Pi + (s, a)) \geq (1 - \epsilon)V^*$ **then**
        $\Pi' \leftarrow$ getOptimal $(\Pi + (s, a), i + 1, \epsilon)$
        **if** $g(\Pi') > g(\Pi_o)$ **then**
            $\Pi_o \leftarrow \Pi'$
        **end**
    **end**
**end**
**return** $\Pi_o$

---

The algorithm presented in Table 1 is a one-sided recursive depth-first-search algorithm that searches in the space of plausible non-deterministic policies to maximize a function $g(\Pi)$. Here we assume that there is an ordering on the set of state-action pairs $\{p_i\} =$ $\{(s_j, a_k)\}$. This ordering can be chosen according to some heuristic along with a mechanism to cut down some parts of the search space. $V^*$ is the optimal value function and the function $V$ returns the value of the non-deterministic policy that can be calculated by solving the corresponding evaluation MDP.

We should make a call to the above function passing in the conservative policy $\Pi_m$ and starting from the first state-action pair: $getOptimal(\Pi_m, 0, \epsilon)$.

The asymptotic running time of the above algorithm is $O((|S||A|)^d(t_m + t_g))$, where $d$ is the maximum size of an $\epsilon$-optimal policy minus the size of the conservative policy, $t_m$ is the time to solve the original MDP (polynomial in the relevant parameters), and $t_g$ is the time to calculate the function $g$. Although the worst-case running time is still exponential in the number of state-action pairs, the run-time is much less when the search space is sufficiently small. The $|A|$ term is due to the fact that we check all possible augmentations for each state. Note that this algorithm searches in the space of all $\epsilon$-optimal policies rather than only the non-augmentable ones. If we set the function $g(\Pi) = |\Pi|$, then the algorithm will return the biggest non-augmentable $\epsilon$-optimal policy.

This search can be further improved by using heuristics to order the state-action pairs and prune the search. One can also start the search from any other policy rather than the conservative policy. This can be potentially useful if we have further constraints on the problem.

### 4.3.3 Directed Acyclic Transition Graphs

One way to narrow down the search is to only add the action that has the maximum value for any state $s$, and ignore the rest of actions if adding the top action will result in values out of the $\epsilon$-optimality bound:

$$\Pi' = \Pi + \left(s, \operatorname*{argmax}_{a \notin \Pi(s)} Q^{\Pi}(s, a)\right). \tag{33}$$

The modified algorithm will be as follows:

Table 2: Modified heuristic search algorithm with augmentation rule of Eqn 33.

---

**Function** $getOptimal(\Pi, \epsilon)$
$\Pi_o \leftarrow \Pi$
**for** $s \in S$ *where* $\Pi(s) \neq A(s)$ **do**
    $a \leftarrow \operatorname{argmax}_{a \notin \Pi(s)} Q^{\Pi}(s, a)$
    **if** $V(\Pi + (s, a)) \geq (1 - \epsilon)V^*$ **then**
        $\Pi' \leftarrow$ getOptimal $(\Pi + (s, a), \epsilon)$
        **if** $g(\Pi') > g(\Pi_o)$ **then**
            $\Pi_o \leftarrow \Pi'$
        **end**
    **end**
**end**
**return** $\Pi_o$

---

The algorithm in Table 2 leads to a running time of $O(|S|^d(t_m + t_g))$. However this does not guarantee that we see all non-augmentable policies. This is due to the fact that after adding an action, the order of values might change. If the transition structure of the MDP contains no loop with non-zero probability (transition graph is directed acyclic, i.e. DAG), then this heuristic will produce the optimal result while cutting down the search time.

**Theorem 4.** *For MDPs with DAG transition structure, the algorithm of Table 2 will generate all non-augmentable $\epsilon$-optimal policies that would be generated with a full search.*

*Proof.* To prove this, first notice that we can sort the DAG with a topological sort. Therefore, we can arrange the states in levels, having each state only make transitions to states at a future level. It is easy to see that adding actions to a state for a non-deterministic policy can only change the worst-case value of past levels. It will not have any effect on the $Q$-values at the current level or any future level.

Now given any non-augmentable $\epsilon$-optimal policy generated with a full search, there is a sequence of augmentations that generated that policy. Any permutation of that sequence would create the same policy and all the intermediate polices are $\epsilon$-optimal. Now we rearrange that sequence such that we add actions in the reverse order of the level. By the point mentioned above, the $Q$-value of actions at the point where they are being added will not change until the target policy is realized. Therefore all the actions with $Q$-values above the minimum value must be in the policy, or otherwise we can add them, which conflicts with the target policy being non-augmentable. Since all the actions above a certain $Q$-value must be added, we can add them in order. Therefore the target policy can be realized with the rule of Eqn 33. □

When the transition structure is not a DAG, one might do a partial evaluation of the augmented policy to approximate the value after adding the actions, possibly by doing a few backups rather than using the original $Q$-values. This offers the possibility of trading-off computation time for better solutions.

## 5. Empirical Results

To evaluate our framework and proposed algorithms, we first test both the MIP and search formulations on MDPs created randomly, and then test the search algorithm on a real-world treatment design scenario. Finally, we conduct an experiment on a computer-aided web navigation task with human subjects to assess the usefulness of non-deterministic policies in assisting human decision-making.

### 5.1 Random MDPs

In the first experiment, we aim to study how non-deterministic policies change with the value of $\epsilon$ and how the two algorithms compare in terms of running time. To begin, we generated random MDPs with 5 states and 4 actions. The transitions are deterministic (chosen uniformly at random) and the rewards are random values between 0 and 1, except for one of the states with reward 10 for one of its actions; $\gamma$ was set to 0.95. The MIP method was implemented with MATLAB and CPLEX.

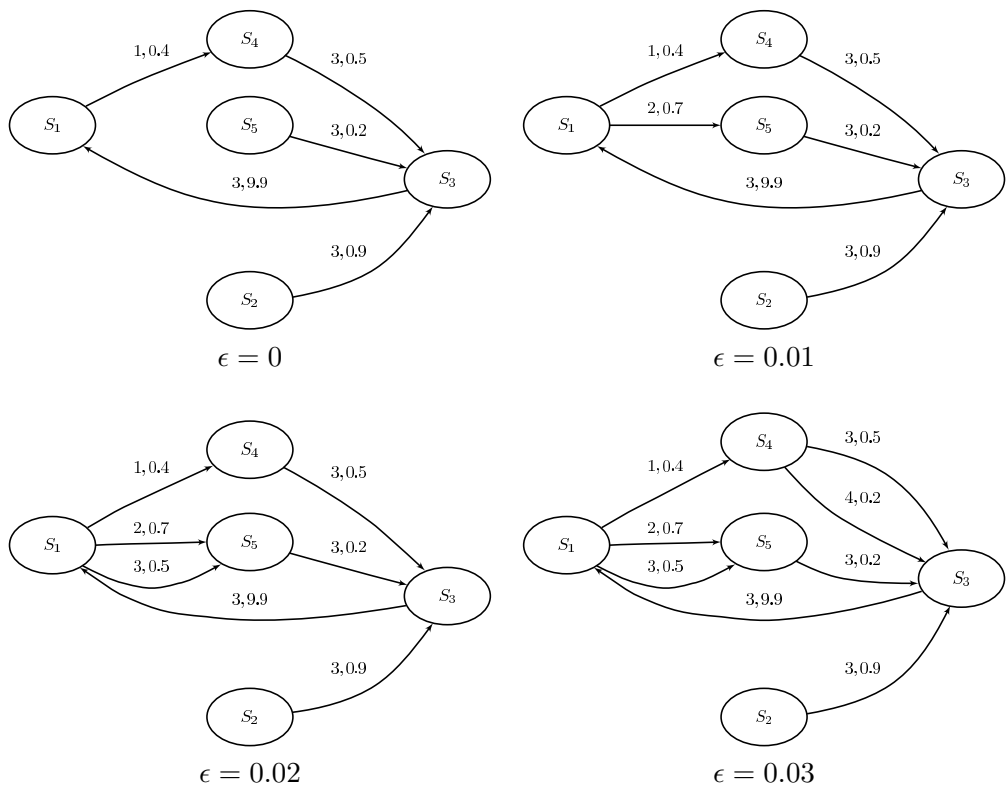

Figure 5: MIP solution for different values of $\epsilon \in \{0, 0.01, 0.02, 0.03\}$. The labels on the edges are action indices, followed by the corresponding immediate rewards.

Figure 5 shows the solution to the MIP defined in Eqn 32 for a particular randomly generated MDP. We see that the size of the non-deterministic policy increases as the performance threshold is relaxed. We can see that even with small values for $\epsilon$ there are several actions included in the policy for each state. This is of course a result of the $Q$-values being close to each other. Such property is typical in many medical scenarios where different treatments provide only slightly different results.

To compare the running time of the MIP solver and the search algorithm, we constructed random MDPs as described above with more state-action pairs. Figure 6 shows the running time averaged over 20 different random MDPs with 5 states, assuming $\epsilon = 0.01$ (which allows several solutions). As expected, both algorithms have a running time exponential in the number of state-action pairs (note the exponential scale on the time axis). The running time of the search algorithm has a bigger constant factor (possibly due to our naive implementation), but has a smaller exponent base, which results in a faster asymptotic running time. Even with the exponential running time, one can still use the search algorithm to solve problems with a few hundred state-action pairs. This is more than sufficient for many practical domains, including real-world medical decision scenarios as shown in the next section.

To observe the effect of the choice of $\epsilon$ on the running time our algorithms, we fix the size of random MDPs to have 7 states and 5 actions at each state, and then change the value of $\epsilon$ and measure the running time of the algorithms over 100 trials. Figure 7 shows the average running time of both algorithms with different values for $\epsilon$. As expected, the search algorithm will go deeper in the search tree as the optimality threshold is relaxed and its running time will thus increase. The running time of the MIP method, on the other hand, remains relativity constant as it exhaustively searches in the space of all possible non-deterministic policies. These results are representative of the relative behaviour of the two approaches over a range of problems.

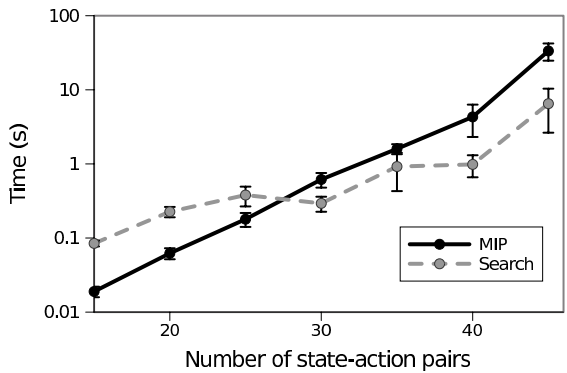

Figure 6: Running time of MIP and the search algorithm as a function of the number of state-action pairs with $\epsilon = 0.01$.

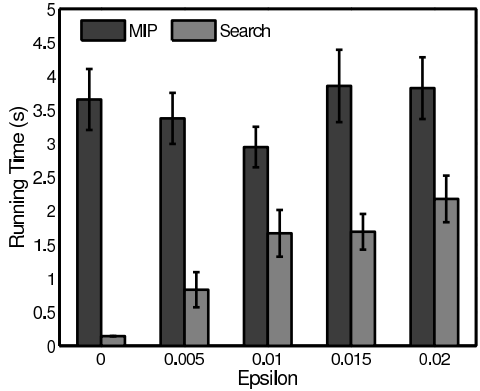

Figure 7: Running time of MIP and the search algorithm as a function of $\epsilon$, with 7 states and 5 actions. Many of the actions are included in the policy with $\epsilon = 0.02$.

## 5.2 Medical Decision-making

To demonstrate how non-deterministic policies can be used and presented in a medical domain, we tested the full search algorithm on an MDP constructed for a medical decision-making task involving real patient data. The data was collected as part of a large (4000+ patients) multi-step randomized clinical trial, designed to investigate the comparative effectiveness of different treatments provided sequentially for patients suffering from depression (Fava et al., 2003). The goal is to find a treatment plan that maximizes the chance of remission. The dataset includes a large number of measured outcomes. For the current experiment, we focus on a numerical score called the Quick Inventory of Depressive Symptomatology (QIDS), which was used in the study to assess levels of depression (including when patients achieved remission). For the purposes of our experiment, we discretize the QIDS scores (which range from 5 to 27) uniformly into quartiles, and assume that this, along with the treatment step (up to 4 steps were allowed), completely describe the patient's *state*. Note that the underlying transition graph can be treated as a DAG, as the study is limited to four steps of treatment and action choices change between these steps. There are 19 actions (treatments) in total. A reward of 1 is given if the patient achieves remission (at any step) and a reward of 0 is given otherwise. The transition and reward models were estimated empirically from the medical database using a frequentist approach.

Table 3: Policy and running time of the full search algorithm on the medical problem.

| | $\epsilon = 0.02$ | $\epsilon = 0.015$ | $\epsilon = 0.01$ | $\epsilon = 0$ |
|---|---|---|---|---|
| Time (seconds) | 118.7 | 12.3 | 3.5 | 1.4 |
| $5 < QIDS < 9$ | CT<br>SER<br>BUP<br>CIT+BUS | CT<br>SER | CT | CT |
| $9 \leq QIDS < 12$ | CIT+BUP<br>CIT+CT | CIT+BUP<br>CIT+CT | CIT+BUP | CIT+BUP |
| $12 \leq QIDS < 16$ | VEN<br>CIT+BUS<br>CT | VEN<br>CIT+BUS | VEN | VEN |
| $16 \leq QIDS \leq 27$ | CT<br>CIT+CT | CT<br>CIT+CT | CT<br>CIT+CT | CT |

Table 3 shows the non-deterministic policy obtained for each state during the second step of the trial (each acronym refers to a specific treatment). This is computed using the search algorithm, assuming different values of $\epsilon$. Although this problem is not tractable with the MIP formulation (304 state-action pairs), a full search in the space of $\epsilon$-optimal policies is still possible. Table 3 also shows the running time of the algorithm, which as expected, increases as we relax the threshold $\epsilon$. Here, we did not use any heuristics. However, as the underlying transition graph is a DAG, we could use the heuristic discussed in the previous section (Eqn 33) to get the same policies even faster.

An interesting question is how to set $\epsilon$ a priori. In practice, a doctor may use the full table as a guideline, using smaller values of $\epsilon$ when he/she wants to rely more on the decision support system, and larger values when relying more on his/her own assessments. We believe this particular presentation of non-deterministic policies could be used and accepted by clinicians, as it is not excessively prescriptive and keeps the physician and the patient in the decision cycle. This is in contrast with the traditional notion of policies in reinforcement learning, which often leaves no place for the physician's intervention.

### 5.3 Human Subject Interaction

Finally, we conduct an experiment to assess the usefulness of non-deterministic policies with human subjects. Ideally, we would like to conduct such experiments in medical settings and with physicians, but such studies are costly and difficult to conduct given that they require participation of many medical professionals. We therefore study non-deterministic policies in an easier domain by constructing a web-based game that can be played by any computer and human (either jointly or separately).

The game is defined as follows. A user is given a target word and is asked to navigate around the pages of Wikipedia and visit pages that contain that target word. The user can click on any word in a page. The system then uses a Google search on the Wiki website with the clicked word and a keyword in the current page (the choice of this keyword is discussed later). It then randomly chooses one of the top eight search results and moves to that page. This process mimics the hyperlink structure of the web (extending over the hyperlink structure of the Wiki to make target words more easily reachable). The user is given ten attempts and is asked to reach as many pages with the target word as possible. A similar game was used in another work to infer the semantic distances between concepts (West, Pineau, & Precup, 2009). Our game, however, is designed in such a way that a computer model can provide results similar to a human player and thus enable us to assess the effectiveness of computer-aided decisions with non-deterministic policies.

We construct this task on the CD version of Wikipedia (Schools-Wikipedia, 2009), which is a structured and manageable version of Wikipedia intended for use in schools. To test our approach we also need to build an MDP model of the task. This is done using empirical data as follows. First, we use Latent Dirichlet Allocation (LDA) using Gibbs sampling (Griffiths & Steyvers, 2004) to divide the pages of Wikipedia into 20 topics. Each topic corresponds to a state in the MDP. The LDA algorithm identifies each topic by a set of keywords that occur more often in pages with that topic. We define each of these sets of keywords to be an action (20 actions totals, each corresponding to 20 keywords). We then randomly navigate around the Wiki using the protocol described above (with a computer player that only clicks LDA keywords) and collect 200,000 transitions. We use the observed data to build the transition and reward model of our MDP (the reward is 1 for each hit and 0 otherwise). The specific choices for the LDA parameter and the number of states and actions in the MDP are made in such a way that the best policy provided by the model has comparable performance with that of the human player.

Using Amazon Mechanical Turk (MTurk, 2010), we consider three experimental conditions with this task. In one experiment, given a target, the computer chooses a word (uniformly at random) from the set of keywords (the action) that comes from the optimal policy on the MDP model. In another experiment, human subjects choose and click the word themselves without any help. Finally, we test the domain with human users while the computer highlights, as hints, the words that come from the non-deterministic policy with $\epsilon = 0.1$. We record the time taken during the process and number of times the target word is observed (number of hits). Table 4 summarizes the average outcomes of this experiment for four of the target words (we used seven target words, but could not collect enough data for all of them). We also include the p-value for the t-test comparing the results for human agents with and without the hints. The computer score is averaged over 1000 runs.

Table 4: Comparison of different agents in the web navigation task. The t-test is between the number of hits for a human player that uses hints and one that does not.

| Target | Computer | Human | Human with hint | t-Test |
|--------|----------|-------|-----------------|--------|
| *Marriage* | 1.88 hits<br><br>(1000 runs) | 1.94 hits<br>103 seconds<br>(86 subjects) | 2.63 hits<br>93 seconds<br>(86 subjects) | 0.012 |
| *Military* | 4.72 hits<br><br>(1000 runs) | 4.86 hits<br>91 seconds<br>(67 subjects) | 5.61 hits<br>84 seconds<br>(97 subjects) | 0.049 |
| *Book* | 3.77 hits<br><br>(1000 runs) | 3.67 hits<br>85 seconds<br>(98 subjects) | 4.39 hits<br>89 seconds<br>(83 subjects) | 0.014 |
| *Animal* | 2.50 hits<br><br>(1000 runs) | 3.18 hits<br>96 seconds<br>(92 subjects) | 3.42 hits<br>85 seconds<br>(123 subjects) | 0.46 |

For the first three target words, where the performance of the computer agent is close to a human user, we observe that providing hints to the user results in a statistically significant increase in the number of hits. In fact we see that the computer-aided human outperforms both the computer and human agents. This shows that non-deterministic policies can provide the means to inject human domain knowledge to computer models in such a way that the final outcome is superior to the decision-making solely performed by one party. For the last word, the computer model is working poorly, judging by its low hit rate. Thus, it is not surprising to see that the hints do not provide much help to the human agent in this case (as seen by the non-significant p-value). We also observe a general speedup (for three of the targets) in the time taken by the agent to choose and click the words, which further shows the usefulness of non-deterministic policies in accelerating the human subjects' decision-making process.

## 6. Discussion

This paper introduces the new concept of non-deterministic policies and their potential use in decision support systems based on Markovian processes. In this context, we investigate how the assumption that a decision-making system should return a single optimal action can be relaxed, to instead return a set of near-optimal actions.

Non-deterministic policies are inherently different from stochastic policies. Stochastic policies assume a randomized action selection strategy with some specific probabilities, whereas non-deterministic policies do not impose such constraint. We can thus use best-case and worst-case analysis with non-deterministic policies to highlight different scenarios with the human user.

The benefits of non-deterministic policies for sequential decision-making are two-fold. First, when we have several actions for which the difference in performance are negligible, we can report all those actions as near-optimal options. For instance, in a medical setting, the difference between the outcome of two treatment options might not be "medically significant". In that case, it may be beneficial to provide all the near-optimal options. This makes the system more robust and user-friendly. In the medical decision-making process, for instance, the physician can make the final decision among the near-optimal options based on side effects burden, patient's preferences, expense, or any other criteria that is not captured by the model used in the decision support system. The key constraint, however, is to make sure that regardless of the final choice of actions, the performance of the executed policy is always bounded near the optimal. In our framework, this property is maintained by an $\epsilon$-optimality guarantee on the worst-case scenario.

Another potential use of the non-deterministic action sets in Markovian decision processes is to capture uncertainties in the optimality of actions. Often times, the amount of data from which models are constructed is not sufficient to clearly identify a single optimal action. If we are forced to chose only one action as the optimal one, we might have a high chance of making the wrong decision. However, if we are given the chance to provide a set of possibly-optimal actions, then we can ensure we include all the promising options while cutting off the obviously bad ones. In this setting, the task is to trim the action set as much as possible while providing the guarantee that the optimal action is still among the top few possible options.

To solve the first problem, this paper introduces two algorithms to find flexible near-optimal policies. First we derive an exact solution with a MIP formulation to find a maximal $\epsilon$-optimal policy. The MIP solution is, however, computationally expensive and does not scale to large domains. We then describe a search algorithm to solve the same problem with less computational cost. This algorithm is fast enough to be applied to real world medical domains. We also show how to use heuristics in the search algorithm to find the solution for DAG structures even faster. The heuristic search can also provide approximate solutions in the general case.

Another way to scale the problem to larger domains is to approximate the solution to the MIP program by relaxing some of the constraints. One can relax the constraints to allow non-integral solutions and penalize the objective for values away from 0 and 1. The study of such approximation methods remains an interesting direction of future work.

The idea of non-deterministic policies introduces a wide range of new problems and research topics. In Section 4, we discuss the idea of near optimal non-deterministic policies and address the problem of finding the one with the largest action set. As mentioned, there are other optimization criteria that might be useful with decision support systems. These include maximizing the decision margin (the margin between the worst selected action and the best one not selected), or alternatively minimizing the uncertainty of a wrong selection. Formalizing these problems into a MIP formulation, or incorporating them into a heuristic search, might prove to be useful.

As evidenced by our human interaction experiments, non-deterministic policies can substantially improve the outcome of planning and decision-making tasks in which a human user is assisted by a robust computer-generated plan. Allowing several suggestions at each step provides an effective way of incorporating domain knowledge from the human side of the decision-making process. In medical domains where the physician's domain knowledge is often hard to capture in a computer model, a collaborative model of decision-making such as non-deterministic policies could offer a powerful framework for selecting effective, and clinically acceptable, treatment strategies.

## Acknowledgments

The authors wish to thank A. John Rush (Duke-NUS Graduate Medical School), Susan A. Murphy (University of Michigan), and Doina Precup (McGill University) for helpful discussions regarding this work. Funding was provided by the National Institutes of Health (grant R21 DA019800) and the NSERC Discovery Grant program.